\documentclass[letterpaper]{article} 
\usepackage{aaai2026}  
\usepackage{times}  
\usepackage{helvet}  
\usepackage{courier}  
\usepackage[hyphens]{url}  
\usepackage{graphicx} 
\usepackage{natbib}  
\usepackage{caption} 
\usepackage[utf8]{inputenc} 
\usepackage[T1]{fontenc}    
\usepackage{url}            
\usepackage{booktabs}       
\usepackage{amsfonts}       
\usepackage{nicefrac}       
\usepackage{microtype}      
\usepackage{graphicx}
\usepackage{supertabular}
\usepackage{bbding}
\usepackage{subfigure}
\usepackage{amsmath}
\usepackage{multirow}
\usepackage{makecell}
\usepackage{colortbl}
\usepackage{tabularx}
\usepackage{amssymb}
\usepackage{caption}
\usepackage{algorithm}
\usepackage{algorithmic}
\usepackage{booktabs}
\usepackage[many]{tcolorbox}
\usepackage{longtable} 
\usepackage{titletoc}

\usepackage{newfloat}
\usepackage{listings}
\DeclareCaptionStyle{ruled}{labelfont=normalfont,labelsep=colon,strut=off} 
\floatstyle{ruled}
\newfloat{listing}{tb}{lst}{}
\floatname{listing}{Listing}
\title{AFTER: Mitigating the Object Hallucination of LVLM via Adaptive Factual-Guided Activation Editing}
\author{
    Tianbo Wang\textsuperscript{\rm 1, \rm 3}
    Yuqing Ma\textsuperscript{\rm 2, \rm 3},
    Kewei Liao\textsuperscript{\rm 1, \rm 3},
    Zhange Zhang\textsuperscript{\rm 2, \rm 3},
    Simin Li\textsuperscript{\rm 1, \rm 3},
    Jinyang Guo\textsuperscript{\rm 2, \rm 3},
    Xianglong Liu\textsuperscript{\rm 1, \rm 3}
}
\affiliations{
    \textsuperscript{\rm 1}School of Computer Science and Engineering, Beihang University\\
    \textsuperscript{\rm 2}Institute of Artificial Intelligence, Beihang University\\
    \textsuperscript{\rm 3}State Key Laboratory of Complex \& Critical Software Environment\\

    tianbowang@buaa.edu.cn
}

\usepackage{bibentry}

\begin{document}

\maketitle

\begin{abstract}
Large Vision-Language Models (LVLMs) have achieved substantial progress in cross-modal tasks. However, due to language bias, LVLMs are susceptible to object hallucination, which can be primarily divided into category, attribute, and relation hallucination, significantly impeding the trustworthy AI applications. Editing the internal activations of LVLMs has shown promising effectiveness in mitigating hallucinations with minimal cost. 
However, previous editing approaches neglect the positive guidance offered by factual textual semantics, thereby struggling to explicitly mitigate language bias. 
To address these issues, we propose \textbf{A}daptive \textbf{F}actual-guided Visual-\textbf{T}extual \textbf{E}diting fo\textbf{R} hallucination mitigation (AFTER), which comprises Factual-Augmented Activation Steering (FAS) and Query-Adaptive Offset Optimization (QAO), to adaptively guide the original biased activations towards factual semantics. Specifically, FAS is proposed to provide factual and general guidance for activation editing, thereby explicitly modeling the precise visual-textual associations.
Subsequently, QAO introduces a query-aware offset estimator to establish query-specific editing from the general steering vector, enhancing the diversity and granularity of editing.
Extensive experiments on standard hallucination benchmarks across three widely adopted LVLMs validate the efficacy of the proposed AFTER, notably achieving up to a 16.3\% reduction of hallucination over baseline on the AMBER benchmark. Our code and data will be released for reproducibility.
\end{abstract}

\section{Introduction}\label{sec:intro}
Building upon the foundation of Large Language Models (LLMs), Large Vision-Language Models (LVLMs) have made substantial advancements in cross-modal understanding and generation \cite{qwen, mplug}. 
However, LVLMs continue to grapple with a significant challenge known as \emph{object hallucination} \cite{survey1,survey2}, which refers to discrepancies between the factual visual objects and the model-generated response. This issue severely impedes the trustworthiness of LVLMs in real-world applications \cite{yan2024med,xie2025vlms}.

\begin{figure}[!t]
\centering
\includegraphics[width=1\linewidth]{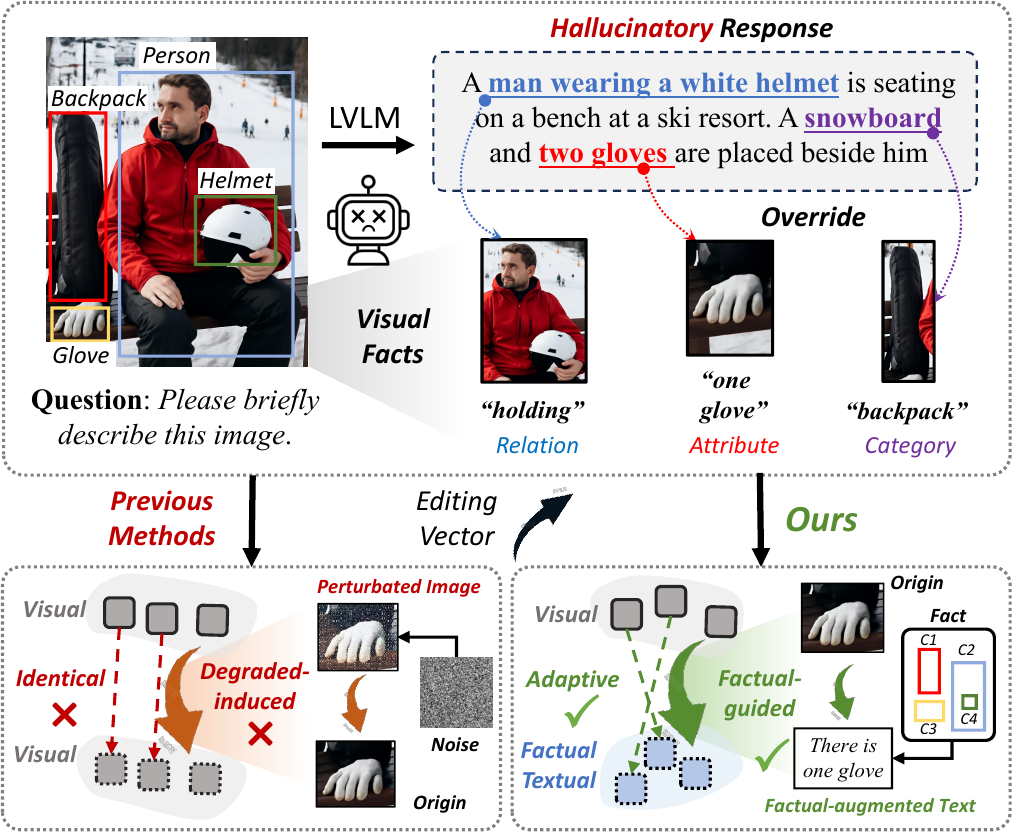}
\caption{The above figure demonstrates the three types of hallucinations (category, attribute, and relation) caused by language bias. The below figure shows the comparisons between previous activation editing methods and AFTER. 
}
\label{fig:intro}
\end{figure}

Existing studies have demonstrated that one primary cause of hallucination is the language bias \cite{survey1, devils,vcd,tuning}, which leads LVLM to prioritize textual knowledge over the external visual inputs. 
As illustrated in Figure \ref{fig:intro}, language bias empirically results in three primary types of hallucination \cite{survey1, survey2}: 
(1) \emph{Category Hallucination}: The object category ``backpack'' is mistakenly identified as a ``snowboard'' due to the language prior associating skiing with snowboards \cite{niu2021counterfactual}. (2) \emph{Attribute Hallucination}: The incorrect object attribute (\emph{e.g.} counting) of gloves arises from the prior that gloves typically appear in pairs \cite{niu2021counterfactual, agrawal2018don}. (3) \emph{Relation Hallucination}: The frequent prior ``man wearing a helmet'' overrides the object relation fact ``man holding a helmet'' \cite{agrawal2018don}.
Although existing hallucination mitigation methods, \emph{e.g.} training-based \cite{dpo,wang2024mitigating} and inference-time \cite{halc, kimvacode}, have gained notable success, their practical applications are constrained by either excessive training burden or multi-round inference costs \cite{ict}.

Recently, inference-time activation editing techniques \cite{iti, trfr, sea, truthx} have shown promise in addressing hallucinations in LVLMs \cite{ict,vti}. Through employing carefully designed editing vectors, these techniques can directly optimize LVLMs' behavior by editing the hallucinatory internal activation with minimal inference costs. 
For instance, VTI \cite{vti} constructs the vector by contrasting stable visual features (averaged from multiple perturbed images) with the original ones, then applies interventions in the visual encoder to enhance activation stability. ICT \cite{ict} generates globally noisy and locally blurred images as untrusted semantics, 
which are used to calculate separate editing vectors to improve the comprehension of image information and object details in LVLMs, respectively.

However, although prior methods intentionally degrade visual semantics (\emph{e.g.} injecting perturbations into images) to steer activations within the visual space, they overlook the positive guidance offered by factual textual semantics.
As a result, these methods fail to capture diverse visual-textual associations, limiting their ability to explicitly mitigate language bias. Specifically, the factual information embedded in the image's ground-truth annotations cannot be textualized by existing methods to construct positive steering directions, thereby failing to tackle visual-textual disparity \cite{hacl, farlhf}.
Additionally, the diverse query-emphasized objects exhibit distinct visual-textual associations with specific offsets from the general one, which existing identical steering vectors cannot accommodate.

Therefore, we propose \textbf{A}daptive \textbf{F}actual-Guided Visual-\textbf{T}extual \textbf{E}diting fo\textbf{R} hallucination mitigation (\textbf{AFTER}), which comprises \emph{\textbf{F}actual-Augmented \textbf{A}ctivation \textbf{S}teering (FAS)} and  \emph{\textbf{Q}uery-\textbf{A}daptive Offset \textbf{O}ptimization (QAO)},
to adaptively steer original activation toward factual-augmented textual semantics for language bias alleviation. 
FAS first leverages factual information to provide positive and explicit textual guidance for visual-textual activation editing. It innovatively transforms ground-truth annotations into textual category, attribute, and relation facts, thereby generating trusted text-query samples that are resistant to language bias. Subsequently, FAS can derive a general and positive visual-textual steering direction by contrasting trusted textual activations with original activations, thereby effectively guiding the activations to tackle visual-textual disparity.
To further promote editing diversity, QAO introduces a query-aware offset estimator to assess distinct deviations from the general steering vector, therefore establishing query-specific visual-textual associations.  
QAO specifically evaluates the overlap between query-referenced objects and entire category facts to generate query-specific offsets.
This guides the estimator to adaptively steer LVLMs towards prioritizing edited visual semantics, thereby mitigating language bias. 
We summarize our contributions as follows: 
\begin{itemize}
    \item We propose the AFTER, an effective activation editing approach to adaptively steer original activation toward factual-augmented semantics for hallucination mitigation.
    \item We introduce Factual-Augmented Activation Steering (FAS), which leverages factual textual semantics to provide positive guidance for activation editing of LVLM.
    \item We propose Query-Adaptive Offset Optimization (QAO), which further establishes query-specific visual-textual association based on the general vector to promote diversity.
    \item Extensive experiments reveal that our method achieves superior performance with minimal cost, outperforming baselines by up to 16.3\% reduction on AMBER. It also exhibits strong generalizability and proves effective in enhancing common visual-textual capability.
\end{itemize}

\section{Related Works}\label{sec:related}

\subsection{Large Vision-Language Models}\label{sec:related_LVLM}
Building on the successful application of Large Language Models (LLMs), Large Vision-Language Models (LVLMs) enhance the visual perception of LLMs \cite{llama,vicuna} by integrating a pre-trained visual encoder \cite{clip,eva}, achieving significant performance in diverse vision-language tasks \cite{flickr30k, cococaption, aokvqa, gqa}. To establish the connection between visual and textual representation, LVLMs usually incorporate a learnable interface, which can be broadly classified into query-based and projection-based\cite{survey1,hacl}. Query-based methods, such as InstructBLIP \cite{instructblip}, MiniGPT-4 \cite{minigpt} with Q-Former, utilize a set of learnable query tokens to capture visual signals via cross-attention. Represented by LLaVA \cite{llava} and Shikra \cite{shikra}, projection-based methods utilize a trainable linear projection layer or a Multi-Layer Perceptron (MLP) to transform extracted visual features. In this work, we selected three commonly used LVLMs of LLaVA-v1.5, Shikra, and InstructBLIP to evaluate our approach.
\subsection{Hallucination Mitigation of LVLM}\label{sec:related_Hal}
Current LVLM hallucination mitigation methods fall into training-based and inference-time approaches. Training-based methods retrain LVLMs with high-quality data \cite{tuning,rlhfv,dpo} or new objectives \cite{hacl,hio}, but are time-consuming and resource-intensive. Inference-time methods mitigate hallucinations during generation via specialized decoding \cite{vcd, opera, halc} or iterative corrections \cite{volcano, woodpecker}, but require multiple inference steps that increase inference cost. Currently, several works \cite{vti,ict} have demonstrated that directly editing the internal activations of LVLM during inference can mitigate hallucination. 
For example, VTI \cite{vti} constructs a vector by contrasting stable visual features (averaged from perturbed images) with the original ones, then applies interventions in the visual encoder to enhance activation stability. ICT \cite{ict} generates globally noisy and locally blurred images as untrusted semantics, computing separate editing vectors to improve the comprehension of image information and object details in LVLMs.
However, they fail to capture the query-specific visual-textual association, thereby limited to explicitly mitigate language bias.

\begin{figure*}[!t]
\centering
\includegraphics[width=1\linewidth]{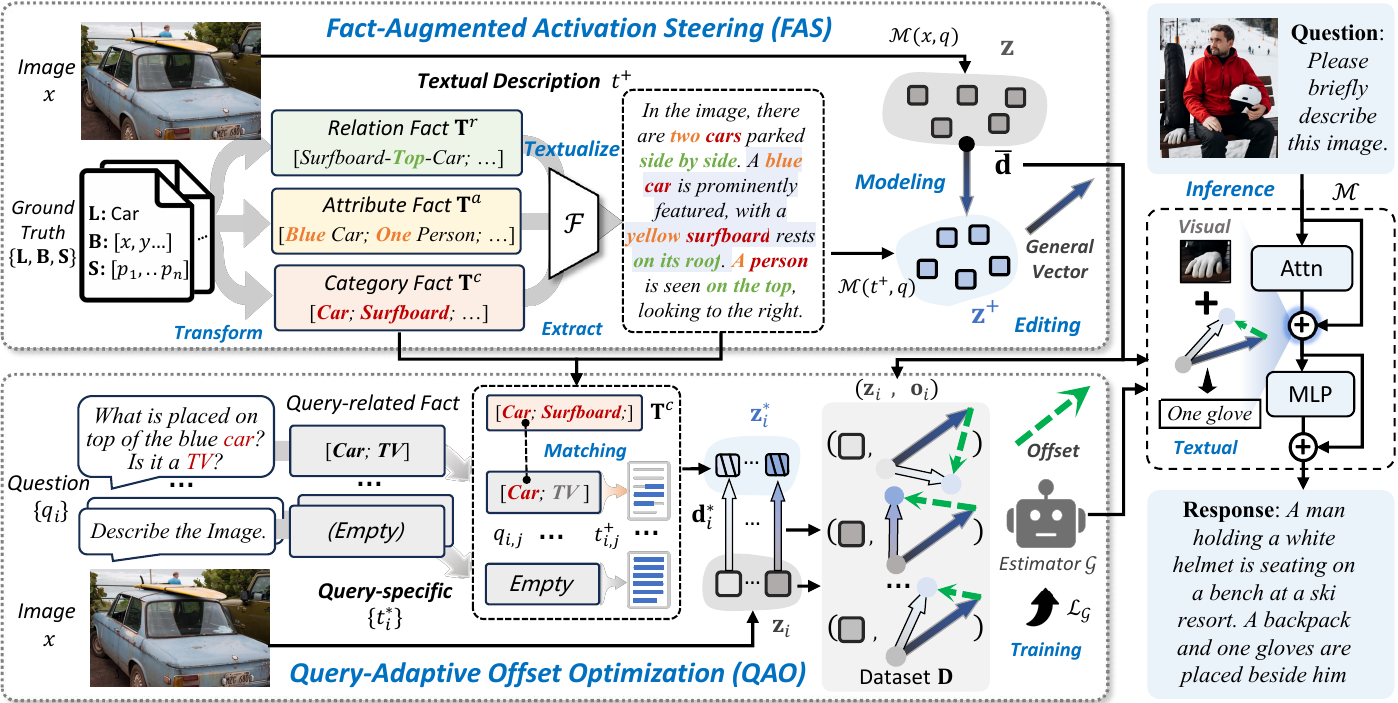}
\caption{An overview of the AFTER. FAS first establishes the general and positive visual-textual editing direction with the guidance of facts. QAO then achieves precise query-adaptive editing by training a query-aware offset estimator, thereby explicitly mitigating language bias. }
\label{fig:method}
\end{figure*}

\section{Methodology}\label{sec:method}
To effectively reduce query-specific language bias, we introduce Adaptive Factual-guided Visual-Textual Editing foR hallucination mitigation (AFTER). AFTER initially leverages Factual-Augmented Activation Steering (FAS) to establish the general and truthful visual-textual editing direction, thereby steering original hallucinatory activation toward factual-guided textual semantics. Subsequently, Query-Adaptive Offset Optimization (QAO) is introduced to generate necessary offset on the general vector, enabling adaptive and precise editing for distinct queries.
In this section, we first present preliminary in Section \ref{sec:method_preliminary}, and elaborate on FAS in Section \ref{sec:method_inno1} and QAO in Section \ref{sec:method_inno2}.
\subsection{Preliminary}\label{sec:method_preliminary}

Given an LVLM $\mathcal{M}$ encoded with rich pretrained language knowledge, the model can process a query composed of an image-question pair $\langle x,q\rangle$, and generate an answer $y=\mathcal{M}(x,q)$. During forward of $\mathcal{M}$, the image-question pair $\langle x,q\rangle$ is tokenized and subsequently passed through $L$ decoding layers with $H$-head self-attention, yielding the hidden states at each layer as $\mathbf{h}^l$:
\begin{equation}
\mathbf{h}^{l+1}=\mathbf{h}^l+\text{Concat}_{k=1}^H(\mathbf{z}^{l,k})\cdot W_o^l,
\end{equation}
where $\mathbf{z}^{l,k}=\text{Attn}^{l,k}(\mathbf{h}^{l})$ denotes the internal activation after self-attention operation of the $k$-th head at the $l$-th layer, $W_o^l$ is an output projection matrix. However, $\mathcal{M}$ tends to prioritize textual knowledge over the external visual input $x$ due to language bias, rendering the generated answer $y$ to be hallucinatory. Therefore, sparse interventions on the internal activations have been designed by activation editing to guide the model toward producing non-hallucinatory outputs.

Typically, these methods first construct steering vector $\bar{\mathbf{d}} = \sum_{\mathbf{X}}(\mathbf{z}^+-\mathbf{z}^-)/|\mathbf{X}|$ by averaging the differences between trusted visual activation $\mathbf{z}^+$ \footnote{Due to the identical operation, we omit the layer $l$ and head $k$ indices in the upper right corner for all activation symbols in following Sections to simplify the notation.} and untrusted visual activation $\mathbf{z}^-$ across image set $\mathbf{X}$. The editing vector $\bar{\mathbf{d}}$ is then applied to the internal activation during inference as follows:
\begin{equation}
\mathbf{h}^{l+1}=\mathbf{h}^l+\text{Concat}_{k=1}^H(\mathbf{z}^{l,k}+\alpha\cdot \bar{\mathbf{d}})\cdot W_o^l,
\end{equation}
where $\alpha$ denotes the editing intensity. 
However, previous methods typically degrade image $x$ to obtain trusted activation $\mathbf{z}^+$ and untrusted activation $\mathbf{z}^-$, failing to establish factual steering guidance. In contrast, our FAS in Section \ref{sec:method_inno1} augment $x$ with abundant facts to generate factual textual description $t^+$, thereby providing positive guidance by extracting $\mathbf{z}^+$ from factual $(t^+,q)$ and $\mathbf{z}^-$ from original $(x,q)$. Additionally, prior researches ignore query-specific visual-textual associations and employ identical averaged vectors $\bar{\mathbf{d}}$ for editing. Our QAO in Section \ref{sec:method_inno2} specially estimates the query-specific offset $\mathbf{o}_i$ based on the averaged vector $\bar{\mathbf{d}}$, realizing query-adaptive factual-guided activation editing.


\subsection{Factual-Augmented Activation Steering}\label{sec:method_inno1}
To fully exploit factual textual semantics for positive editing guidance, we propose Factual-Augmented Activation Steering (FAS) to directly reduce language bias. 
FAS intuitively treats the original visual information as untrusted semantics, and our fact-augmented textual description as trusted semantics, therefore explicitly constructing reliable visual-textual editing vectors. This enables positive steering of the original hallucinatory activation, thereby preventing the misguidance of language bias.

To facilitate the generation of factual textual descriptions as trusted semantics, we innovatively textualize the ground-truth annotations into category, attribute, and relation facts, thereby effectively mitigating the three types of hallucination. Specifically, we sample an image set $\mathbf{X}$ from the classic COCO \cite{coco} training set, each image $x\in \mathbf{X}$ accompanied by rich ground-truth annotations of core objects. 
The transformations of ground-truth annotations into category fact set $\mathbf{T}^c$, attribute fact set $\mathbf{T}^a$, and relation fact set $\mathbf{T}^r$ are illustrated as follows (Details are presented in Appendix C):
\begin{itemize}
  \item \textbf{Category fact set} $\mathbf{T}^c$: The category facts correspond to the factual description of object categories, which can be generated by directly integrating the category labels $\mathbf{L}$ of all objects.
  \item \textbf{Attribute fact set} $\mathbf{T}^a$: In the attribute fact set $\mathbf{T}^a$, the focused facts primarily include color, shape, and count:
  \begin{itemize}
    \item \textbf{Color}: The color attribute is manually annotated based on pixel-level statistics within the objects. 
    We specifically designate the color with the highest pixel proportion in segmented region as the object's color attribute.
    \item \textbf{Shape}: This attribute refers to the objects' shape (\emph{e.g.} circular, square), which are transformed from the segmentation polygons $\mathbf{S}$ by approximating their contours with polygonal curves and analyzing geometric regularities such as vertex count and angular consistency.
    \item \textbf{Count}: The count attribute denotes the occurrence frequency of a particular category within the image, which can be calculated according to category labels $\mathbf{L}$.
\end{itemize}
  \item \textbf{Relation fact set} $\mathbf{T}^r$: Relation facts can be estimated from the spatial relationships (\emph{e.g.} left, overlapped) between bounding boxes annotations $\mathbf{B}$. This process is achieved by computing the directional offsets between the box centers and spatial proximity according to their IoU score.
\end{itemize}
After accurately extracting the three types of hallucination-related facts, we textualize all the facts into a comprehensive and factual description with the help of existing LVLM:
\begin{equation}
t^+=\mathcal{F}(\text{I}_{\text{fst}};(x,[\mathbf{T}^c,\mathbf{T}^a, \mathbf{T}^r])),
\end{equation}
where $t^+$ denotes the textualized factual description by LVLM $\mathcal{F}$ with instruction $\text{I}_{\text{fst}}$ (shown in Appendix F). It is worth noting that $\mathcal{F}$ is employed solely for integrating discrete facts into coherent textual ground-truth, which is necessary for editing methods \cite{iti}, without providing extra information.
The capabilities of $\mathcal{F}$ are not engaged during the inference of the edited model $\mathcal{M}$, thereby ensuring a fair comparison with other methods.

Subsequently, FAS can construct trusted-untrusted sample pairs $\langle(t^+,q),(x,q)\rangle$ by concatenating trusted textual description $t^+$ and untrusted visual images $x$ with question $q$, facilitating the modeling of positive editing directions.
Specifically, for each image $x$ and corresponding textual description $t^+$, we construct an $n$-question set $\{q_i\}$ associated with diverse object facts, where each question $q_i$ (\emph{e.g., Describe this image.}) has the potential to elicit a hallucinatory response. Subsequently, we combine visual image $x$ and textual description $t^+$ with every generated question, forming $n$ trusted-untrusted sample pairs $\{\langle(t^+,q_i),(x,q_i)\rangle|i\in[1,n]\}$. The samples are then input into LVLM $\mathcal{M}$ to obtain the trusted-untrusted activation pairs $\langle \mathbf{z}_{i}^+, \mathbf{z}_i\rangle$, which represent the factual textual semantics and original hallucinatory semantics perceived by $\mathcal{M}$, respectively. Therefore, we can directly model the general visual-textual steering vector by averaging the computed differences between $\mathbf{z}_i^+$ and $\mathbf{z}_i$ across the whole image set $\mathbf{X}$, which is a common practice for activation editing \cite{ict, destein}:
\begin{equation}
\bar{\mathbf{d}}=\frac{1}{n\cdot |\mathbf{X}|}\sum_{\mathbf{X}}\sum_{i=1}^n(\mathbf{z}_i^+  -\mathbf{z}_i),
\end{equation}
where $\bar{\mathbf{d}}$ denotes the general visual-textual editing vector, $|\mathbf{X}|$ denotes the number of calculated images. Therefore, FAS can explicitly reduce the language bias by applying the general steering vector to perform beneficial editing, thereby mitigating the hallucinatory response.
\begin{table*}[t]
\centering
\small

\renewcommand{\arraystretch}{0.9}
\begin{tabular}{c|c|cc|cccc|ccc}
\toprule
\multirow{2.3}*{\textbf{Models}} & \multirow{2.3}*{\textbf{Methods}} & \multicolumn{2}{c|}{\textbf{POPE}} & \multicolumn{4}{c|}{\textbf{MME}} & \multicolumn{3}{c}{\textbf{AMBER}} \\
& & \raisebox{-0.5ex}{\textbf{\ ACC(↑)\ }} & \raisebox{-0.5ex}{\textbf{\ F1(↑)\ }} & \raisebox{-0.5ex}{\textbf{\ E(↑) \ }} & \raisebox{-0.5ex}{\textbf{\ CT(↑)\ }} & \raisebox{-0.5ex
}{\textbf{\ P(↑) \ }} & \raisebox{-0.5ex}{\textbf{\ CR(↑)\ }} & \raisebox{-0.5ex}{\textbf{\ CHAIR(↓) \ }} & \raisebox{-0.5ex}{\textbf{\ Hal(↓) \ }} & \raisebox{-0.5ex
}{\textbf{\ Cover(↑)\ }}\\
\midrule

\multirow{8}{*}{\makecell{\textbf{LLaVA-} \\ \textbf{v1.5}}} & \textbf{Baseline} & 80.1 & 82.3  & 180.0 & 158.3 & 123.3 & 155.0 & 6.9 & 31.6 & 48.9 \\
 & \textbf{HACL}& 83.5 & 83.0 & 185.0 & \textbf{168.3} & 133.3 & 145.0 & 7.1 & 31.4 & \textbf{49.6} \\
 & \textbf{VCD}& 82.5 & 82.7 & 190.0 & 148.3 & 126.7 & 158.3  & 5.1 & 27.6 & 48.6 \\

 & \textbf{OPERA}& 83.3 & 83.5 & 190.0 & 153.3 & 123.7 & 158.3  & 4.9 & 27.9 & 49.0 \\
 & \textbf{VTI} & 83.2 & 83.4 & 185.0 & 163.3 & 128.3 & 150.0  & 5.1 &  23.7 & 47.8 \\

 & \textbf{ICT}& 83.7 & 83.7 & \textbf{195.0} & 158.3 & 126.7 & 158.3  & 5.4 & 26.6 & 48.8 \\
 \cmidrule(r){2-11}
  & \textbf{\emph{w/o} QAO} & 83.8  & 84.4 & \textbf{195.0}  & 163.3 & 128.3 & 160.0  & 5.2 & 22.3 & 48.6 \\
 
 & \textbf{Ours} & \textbf{85.7}  & \textbf{85.6} & \textbf{195.0}  & 163.3 & \textbf{138.3} & \textbf{165.0}  & \textbf{4.5} & \textbf{20.5} & 48.7 \\
\midrule

\multirow{8}{*}{\makecell{\textbf{Instruct-} \\ \textbf{BLIP}}} & \textbf{Baseline} & 80.3 & 82.0 & 175.0 & 60.0 & 50.0 & 120.0  & 7.4 & 35.4 & 53.5\\
 & \textbf{VCD}& 81.5 & 82.1 & 180.0 & 60.0 & 48.3 & 125.0  & 6.9 & 32.3 & \textbf{53.8} \\

 & \textbf{OPERA}& 82.0 & 82.3 & 180.0 & 65.0 & 58.3 & 128.3  & 6.6 & 31.4 & 53.5\\
 & \textbf{VTI} & 82.3 & 82.7 & 170.0 & 60.0 & 53.3 & 120.0  & 5.3& 26.7 & 53.0\\

 & \textbf{ICT} & 82.6 & 82.9 & 180.0 & 60.0 & 56.7 & 130.0  & 6.2& 30.8 & 53.6\\
\cmidrule(r){2-11}
  & \textbf{\emph{w/o} QAO} & 82.9  & 83.8 & \textbf{185.0} & 65.0 & 53.3 & 128.3  & 5.8 & 28.6 & 53.7 \\
 
 & \textbf{Ours} & \textbf{83.5}  & \textbf{84.2} & \textbf{185.0} & \textbf{70.0} & \textbf{63.3} & \textbf{133.3}  & \textbf{5.2} & \textbf{25.1} & 53.6 \\
\midrule

\multirow{8}{*}{\makecell{\textbf{Shikra}}} & \textbf{Baseline} & 78.9 & 80.3 & 185.0 & 66.7 & 58.3 & 103.3  & 10.9 & 49.5 & 50.7 \\
 & \textbf{VCD}& 80.2 & 81.2 & 185.0 & 86.7 & 60.0 & 96.7  & 9.7 &  46.9 & 50.2\\

 & \textbf{OPERA}& 80.2 & 81.1 & 185.0 & 85.0 & 63.3 & 106.7 & 8.9 & 42.8 & \textbf{51.0} \\
 & \textbf{VTI}  & 80.6 & 81.3 & 185.0 & 83.3 & 55.0 & 101.7  & 7.5 & 38.5 & 48.6 \\

 & \textbf{ICT} & 80.9 & 81.6 & \textbf{190.0} & 95.0 & 61.7 & 103.7  & 8.7 & 42.5 & 50.8 \\
\cmidrule(r){2-11}
 & \textbf{\emph{w/o} QAO} & 81.1  & 81.6 & \textbf{190.0} & 106.7 & \textbf{66.7} & 103.7 & 7.9 & 38.2 & 50.6 \\
 
 & \textbf{Ours} & \textbf{82.5}  & \textbf{82.5} & \textbf{190.0} & \textbf{116.7} & \textbf{66.7} & \textbf{113.3}  & \textbf{6.9}& \textbf{33.2} & 50.4\\
 
\bottomrule
\end{tabular}

\caption{Comparison of AFTER with SOTA methods on POPE, MME, and AMBER. \textbf{\emph{w/o} QAO} denotes our AFTER excluding QAO. The best results are in \textbf{bold}.  Each result is reported under multiple rounds. For POPE, we report the average Accuracy and F1-score across the three datasets (COCO, A-OKVQA, GQA) and three settings (random, popular, and adversarial). The short names ``E'', ``CT'', ``P'', and ``CR'' refer to existence, count, position, and color dimensions in MME, respectively.
}
\label{tab:pope}
\end{table*}

\subsection{Query-Adaptive Offset Optimization}\label{sec:method_inno2}
Distinct visual semantics emphasized by different queries require specialized editing to more precisely reduce language bias. This motivates the need to apply an adaptive offset on the general visual-textual vector, thereby constructing steering vectors tailored to the specific query. To this end, we propose Query-Adaptive Offset Optimization (QAO), which intuitively devises a query-aware offset estimator that fully captures query-relevant visual semantics and estimates the necessary offset accordingly.



To provide a specific data foundation for training the offset estimator, we first generate more detailed textual descriptions 
of query-emphasized visual semantics. Specifically, given image $x$, its textual description $t^+$, and a question $q_i$, we first extract all object categories $\{q_{i,j}\}$ mentioned in $q_i$, which constitute the query-relevant visual details that LVLM is expected to attend to. Therefore, we seek to obtain object-related textual description $t^+_{i,j}$ of each $q_{i,j}$ according to the following principles:
\begin{equation}
    t^+_{i,j}=\begin{cases}\mathcal{F}(\text{I}_\text{qst}; t^+, q_{i,j})&,   q_{i,j} \in \mathbf{T}^c \\  
   \text{``\emph{There is no} [$q_{i,j}$] \emph{in the image.}''}&,  q_{i,j} \notin \mathbf{T}^c
   \end{cases}
\end{equation}
This process means that if the query-related object is present in the image (\emph{i.e.} $q_{i,j} \in \mathbf{T}^c$), we prompt $\mathcal{F}$ with instruction $\text{I}_\text{qst}$ to extract the corresponding sub-description related to $q_{i,j}$ from the whole textual description $t^+$. Otherwise, we explicitly describe that the queried object is not present in the image.
It is noticed that if $q_{i,j}$ does not mention any object (\emph{e.g. Please describe this image.}), the original textual description $t^+$ is retained.
By consolidating all detailed descriptions $t_{i,j}^+$ derived from the object categories, we ultimately obtain the query-focused textual factual semantic $t_i^*=[t_{i,j}^+]_{j=1}^n$.

Upon obtaining the query-emphasized textual description, we are able to construct query-focused trusted-untrusted sample pairs $\langle(t_i^*,q_i),(x,q_i)\rangle$, and extract corresponding trusted-untrusted activation pairs $\langle\mathbf{z}_{i}^*,\mathbf{z}_i\rangle$. The precise query-specific disparity $\tilde{\mathbf{d}}_i = \mathbf{z}_i^*-\mathbf{z}_i$ serves as the optimal editing vector for the current query. Therefore, aiming to estimate the necessary offset needs to be added on the general vector $\bar{\mathbf{d}}$, we construct a training dataset $\mathbf{d}=\{(\mathbf{z}_i,\mathbf{o}_i)|i\in[1,n]\}$, where $\mathbf{o}_i=\tilde{\mathbf{d}}_i-\bar{\mathbf{d}}$ denotes the expected offset. Based on $\mathbf{D}$, we train the offset estimator $\mathcal{G}$ to comprehend the query-focused visual semantics $\mathbf{z}_i$ and estimate the offset $\mathbf{o}_i$ between the query-specific vector $\tilde{\mathbf{d}}_i$ and the general vector $\bar{\mathbf{d}}$. During training, we adopt the Mean-Square Error (MSE) loss to measure the discrepancy between the estimated offset and the expected offset:
\begin{equation}
\mathcal{L}_\mathcal{G}=\frac{1}{n\cdot|\mathbf{X}|}\sum_{\mathbf{X}}\sum_{i=1}^n\|\mathcal{G}(\mathbf{z}_i) - \mathbf{o}_i\|^2.
\end{equation}
Thus, we can obtain the optimized editing vector for steering the query-focused activation towards factual textual semantics. It is worth noting that training $\mathcal{G}$ is highly efficient, as it is both lightweight (single-layer MLP) and does not require fine-tuning of LVLM. More experimental statistics can be seen in Appendix A.2.
Ultimately, we directly apply query-guided editing to the top-$K$ heads most affected by language bias (\emph{i.e.} those exhibiting the largest vector magnitudes). The adaptive visual-textual editing can be formulated as:
\begin{equation}
\mathbf{h}^{l+1}=\mathbf{h}^l+\text{Concat}_{k=1}^H(\mathbf{z}^{l,k}+\alpha\cdot[\mathcal{G}(\mathbf{z}^{l,k})+\bar{\mathbf{d}}])\cdot W_o^l,
\end{equation}
where $\alpha$ denotes the editing intensity. Through query-adaptive factual-guided editing, the LVLM allocates greater attention to the post-edited visual information, thereby mitigating hallucination.
\section{Experiments}\label{sec:exp}

\subsection{Experimental Setup}\label{sec:exp_setup}
\paragraph{Benchmarks and Metrics}
We assess the performance of LVLMs under both discriminative and generative tasks.  \textbf{For \emph{discriminative} task}, we use the widely adopted POPE \cite{pope} and MME \cite{mme} to evaluate diverse types of hallucinations. Following \cite{vcd, ict}, we compare different methods on the POPE task and report the average Accuracy and F1-score across the three datasets (COCO \cite{coco}, A-OKVQA \cite{aokvqa}, GQA \cite{gqa}) and three settings (random, popular, and adversarial). On MME benchmark that evaluates general capabilities as well as object hallucination, we adopt the MME score as the comprehensive metric to provide a quantitative measure. \textbf{For \emph{generative} task}, we employ the generative subset of AMBER \cite{wang2023amber}, which assesses the generative hallucination using metrics CHAIR \cite{chair} and Hal. It also incorporates metric Cover to quantify the comprehensiveness of the response.

\paragraph{Baseline and Comparative Methods}
We choose three commonly-used LVLMs, including LLaVA-v1.5 \cite{llava1.5}, InstructBLIP \cite{instructblip}, and Shikra \cite{shikra} as baselines. 
To evaluate our superiority, we first compare AFTER with existing activation editing methods, \emph{i.e.} VTI \cite{vti} and ICT \cite{ict}. We also consider other typical decoding-based methods that mitigate LVLM hallucination during inference, including VCD \cite{vcd} and OPERA \cite{opera}. Additionally, the training-based method HACL \cite{hacl} is involved for comparison.

\begin{figure}[!t]
    \centering
    \includegraphics[width=1\linewidth]{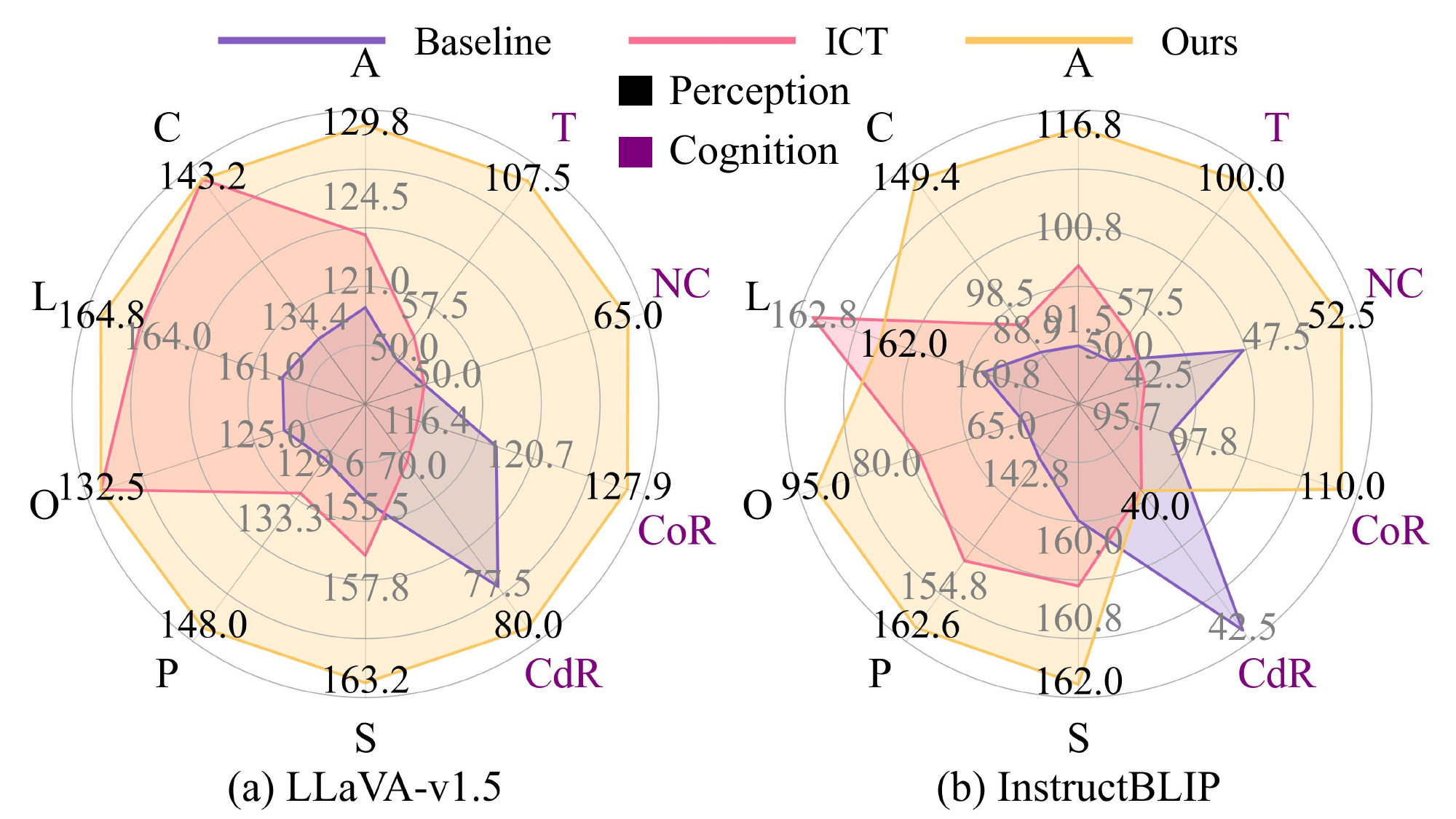}
    \caption{Comparison of AFTER with SOTA editing methods on other perception and cognition capabilities on MME.}
    \label{fig:mme}
\end{figure}

\begin{table}[!t]
    \centering
        \resizebox{1\linewidth}{!}{
        \begin{tabular}{c|c|cc|cc}
            \toprule
                \multirow{2.3}*{\textbf{Models}} & \multirow{2.3}*{\textbf{Methods}} & \multicolumn{2}{c|}{\textbf{COCO $\rightarrow$ GQA}}  & \multicolumn{2}{c}{\textbf{Dis $\rightarrow$ Gen}} \\
                 & & \textbf{ACC} & \textbf{F1} & \textbf{Hal} & \textbf{Cover}\\
            \midrule
                \multirow{2.3}{*}{\makecell{\textbf{LLaVA-} \\ \textbf{v1.5}}} & Baseline & 76.9 & 80.3 & 31.6 & \textbf{48.9}\\
                 & Ours & \textbf{84.6} & \textbf{84.8} & \textbf{22.8} & 48.7 \\
            \midrule
                \multirow{2.3}{*}{\makecell{\textbf{Instruct-} \\ \textbf{BLIP}}} & Baseline & 77.9 & 80.5 & 35.4 & 53.5 \\
                 & Ours & \textbf{81.4} & \textbf{82.4} & \textbf{27.9} & \textbf{53.8} \\
            \midrule
                \multirow{2.3}{*}{\makecell{\textbf{Shikra}}} & Baseline & 78.4 & 80.0 & 49.5 & 50.7 \\
                 & Ours & \textbf{82.3} & \textbf{82.5} & \textbf{38.5} & \textbf{51.2}\\
            \bottomrule
        \end{tabular} 
    } 
    \captionof{table}{Generalization performance of AFTER.
    }
    \label{tab:general}
\end{table}

\paragraph{Implementation Details}
During modeling the visual-textual steering vector, we randomly sample 500 images from the COCO training set, and generate task-specific questions to construct trusted-untrusted sample pairs. We adhere to the experimental setup outlined in \cite{vcd, ict} for fair comparison. Without specifying, the number of edited heads $K$ is set to 64, and the editing strength $\alpha$ is set to 7. More detailed configurations are provided in Appendix C.5. All experiments were conducted on A800.
\subsection{Experimental Results}\label{sec:exp_result}
\paragraph{Hallucination Mitigation Performance}
Table \ref{tab:pope} shows the comparison between AFTER and various hallucination mitigation methods on POPE, MME, and AMBER to illustrate our effectiveness on both discriminative and generative tasks. 

Obviously, our method \textbf{demonstrates \emph{discriminative} advantages} in both POPE and MME benchmarks across three prevailing LVLMs. 
On POPE, we achieve an average improvement of 4.1\% in accuracy and 2.6\% in F1-score over the baselines, surpassing the SOTA editing method ICT by 1.3\% and 0.9\%. 
Additionally, on the hallucination subset of MME \cite{vasparse, ict}, AFTER yields score improvements of 45.0, 46.6, and 73.4 on LLaVA-v1.5, InstructBLIP, and Shikra compared to the vallina LVLM, outperforming all SOTA methods.
This enhancement demonstrates the superiority of the adaptive factual-guided visual-textual editing of AFTER, which effectively avoids the misguidance of language bias by steering original hallucinatory activation towards factual textual semantics.

We also achieve the \textbf{optimal \emph{generative} hallucination mitigation} on AMBER, with an averaged 2.9\% and 12.6\% reduction on CHAIR and Hal metrics over the baselines. When applied to Shikra, we particularly reduce the hallucination by 16.3\%, superior to the suboptimal editing method VTI by 5.3\%. Therefore, without compromising the LVLM's comprehensive understanding of images (negligible change in the Cover metric), AFTER effectively reduces hallucinated objects during generation by leveraging factual visual-textual guidance.
It is noticed that solely deploying the factual-guided vector for editing will bring slightly lower improvement on the three benchmarks. This manifests that query-adaptive editing with the guidance of QAO is also essential for precisely reducing query-specific language bias. 
\par
\begin{table}[!t]
    \centering
        \resizebox{1\linewidth}{!}{
        \begin{tabular}{c|c|cc|cc}
            \toprule
                 \multicolumn{2}{c|}{\multirow{2}{*}{\textbf{Input Semantics}}} & \multicolumn{2}{c|}{\textbf{Direct Input}}  & \multicolumn{2}{c}{\textbf{Steering Vector}} \\
                 \multicolumn{2}{c|}{} & \textbf{ACC} & \textbf{F1} & \textbf{ACC} & \textbf{F1}\\
                \midrule
                \multicolumn{2}{c|}{Image $x$} & 79.2 & 80.9 & - & -\\
                \multicolumn{2}{c|}{Simple Caption $t^{s}$} & 72.5 & 72.8 & 81.4 & 82.2 \\
                \midrule
                
                \multirow{3}{*}{$t^+$} & GPT-4o (200B) & 93.4  & 93.4 & \textbf{85.3} & 84.4 \\
                & GPT-4o-mini (8B) & \textbf{93.9} & \textbf{93.7} & \textbf{85.3} & \textbf{84.5} \\
                & llava-v1.5 (7B) & 92.6 & 92.4 & 85.1 & 84.1 \\
                \bottomrule
        \end{tabular} 
    } 
    \captionof{table}{Comparison of diverse inputs under two strategies. We analyze three variants of $\mathcal{F}$ with varying parameters and architectures for generating factual-augmented text $t^+$.}
    \label{tab:textual_modes}
\end{table}


\paragraph{Foundational Visual-language Performance}
As indicated in Figure \ref{fig:mme}, we also exceed the baseline model and best editing method ICT on almost every dimension that evaluates the general visual perception and cognition capabilities, with an average of 130.7 increased score on three LVLMs. These results indicate that our AFTER not only effectively reduces hallucinations but also enhances general visual capabilities across different models, which benefits from the superiority of steering the visual activation toward factual-guided textual semantics adaptively to alleviate language bias.

\paragraph{Generalization Performance}
We also evaluate the generalizability of AFTER by directly applying the factual visual-textual steering vectors learned from COCO-based discriminative questions to out-of-distribution benchmarks. Specifically, we generalize these vectors on GQA-based POPE evaluation (COCO $\rightarrow$ GQA) and generative AMBER benchmark (Dis $\rightarrow$ Gen) to estimate the generalization performance across visual images and textual questions, respectively. The results in Table \ref{tab:general} demonstrate that AFTER still yields remarkable improvement under different image and question distributions. This indicates that AFTER can achieve general language bias mitigation of LVLMs rather than merely fitting a specific dataset, therefore exhibiting strong generalization.

\subsection{In-depth Analysis}\label{sec:analysis}
\paragraph{Analysis of Factual-augmented Text}\label{sec:analysis_tex}
We employ two strategies: serving as LVLM's input, and steering as trusted activation, to demonstrate the superiority of FAS-derived factual-augmented textual description $t^+$ over simple descriptions $t^{simple}$ (\emph{e.g.} COCO Caption \cite{cococaption}). Table \ref{tab:textual_modes} reveals that simple captions, lacking substantial factual information, perform even 6.7\% worse than visual image $x$ as direct input, and offer marginal guidance in trusted editing. In contrast, our factual textual description encompasses extensive facts, leading to significantly fewer hallucinations than visual images. Furthermore, the visual-textual steering vector derived from FAS more effectively mitigates visual-textual disparity than those from simple captions, demonstrating superior guidance for reducing language bias.

Additionally, the results show that there is minimal performance variation between fact-augmented descriptions $t^+$ generated by LVLMs $\mathcal{F}$ with different parameters and architectures. This demonstrates that the $\mathcal{F}$ employed by FAS is solely utilized for integrating discrete facts into coherent textual ground truth, without distilling new knowledge from $\mathcal{F}$ that would influence the inference of the edited model.

\begin{figure}[!tp]
\centering
\begin{minipage}[t]{0.49\linewidth}
    \centering
    \includegraphics[width=\linewidth]{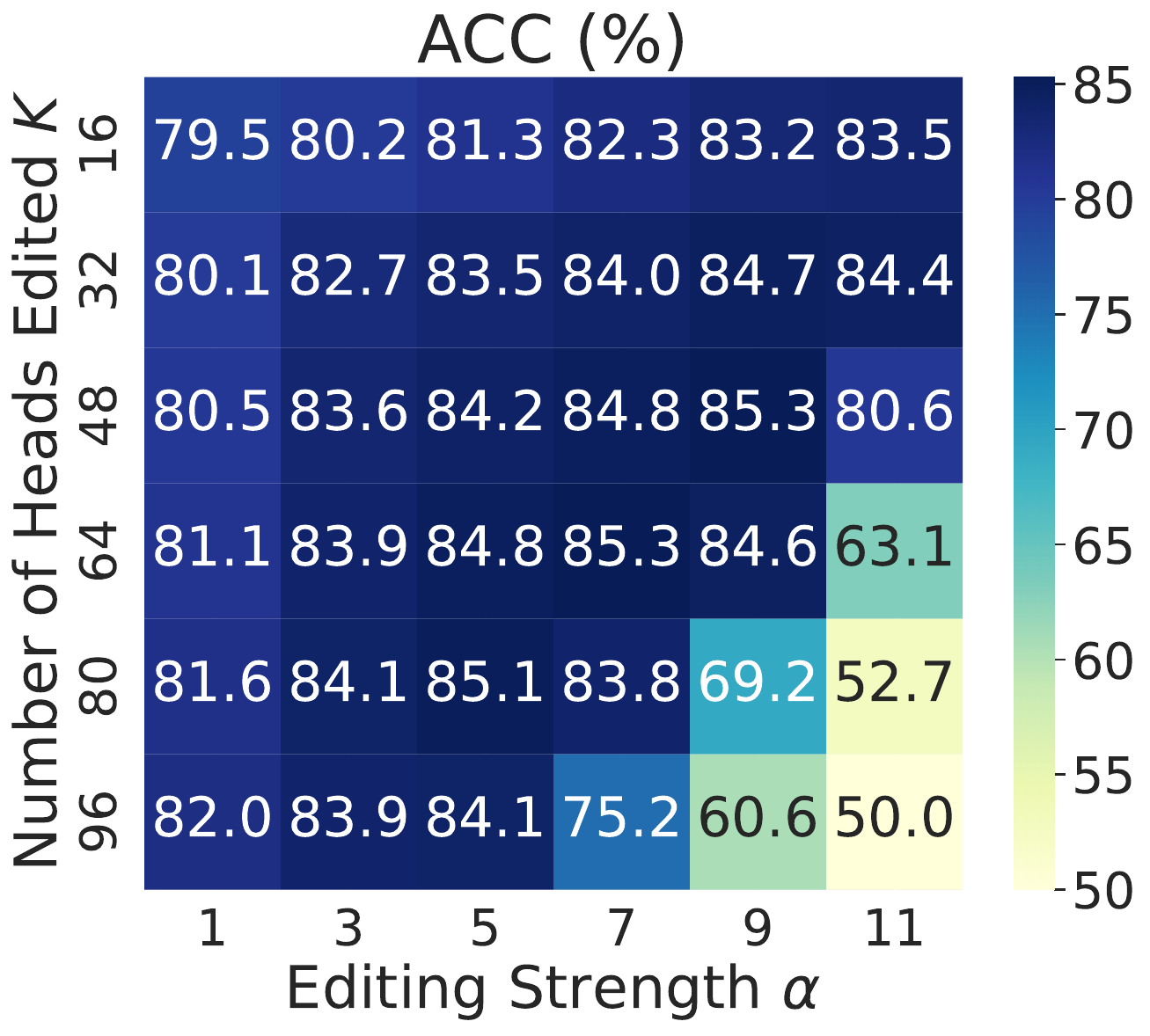}
\end{minipage}%
\hfill
\begin{minipage}[t]{0.49\linewidth}
    \centering
    \includegraphics[width=\linewidth]{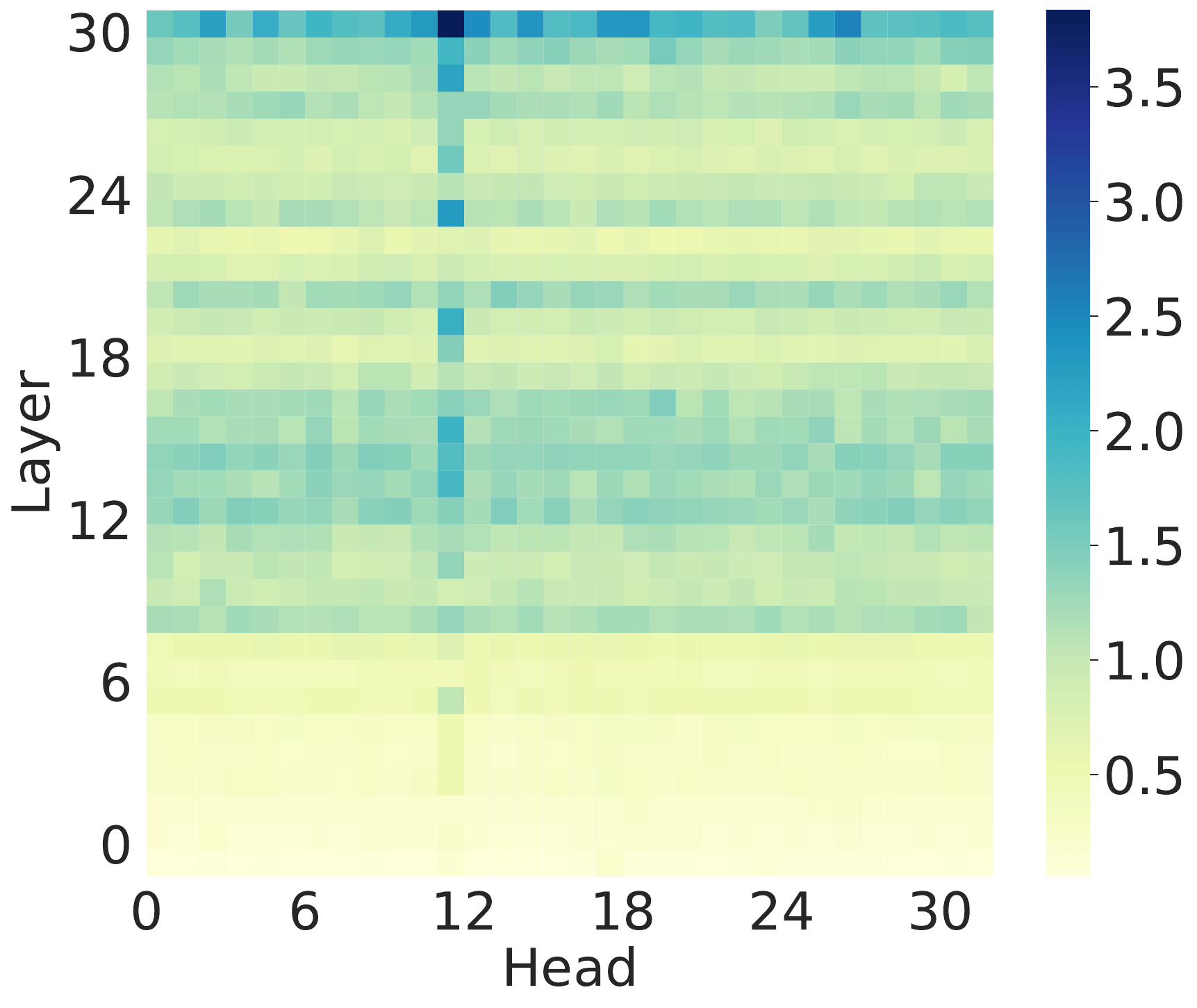}
\end{minipage}
\captionof{figure}{Analysis on LLaVA-v1.5. \textbf{Left}: Ablation of number $K$ and strength $\alpha$. \textbf{Right}: Distribution of vector magnitudes.}
\label{fig:analysis_all_1}
\end{figure}

\paragraph{Analysis of Hyperparameter}\label{sec:analysis_abl}
We analyze two hyperparameters that regulate the editing, \emph{i.e.} the number of edited heads $K$ and editing strength $\alpha$. From the left of Figure \ref{fig:analysis_all_1}, we can observe that both the accuracy and F1 score exhibit an inverted U-shaped curve. The best accuracy (85.3\%) is achieved at $K=64$, $\alpha=7$, while the highest F1 score (84.7\%) appears at $K=64$, $\alpha=9$. These results demonstrate the effectiveness of editing with appropriately calibrated editing strength. The declines under excessive steering reveal a trade-off between truthfulness and helpfulness for editing methods \cite{iti, ict}, providing us with intuitive guidance for editing.

\paragraph{Analysis of Magnitude Distribution}\label{sec:analysis_head}

To investigate the impact of language bias within the LVLM architecture, we analyze the distribution of editing vector magnitudes across all layers and attention heads, as shown on the right of Figure \ref{fig:analysis_all_1}. The results reveal a notable increase in vector magnitudes in the middle layers (layers 9 to 17), which can be attributed to the progressive accumulation of visual information through self-attention \cite{jiang2024devils}. Therefore, language bias significantly interferes with the perception of visual content, resulting in substantial visual-textual disparity. This effect accumulates across subsequent layers and ultimately propagates to the final layer, directly contributing to hallucinatory outputs. Moreover, we observe a particularly pronounced disparity at the 12th head, which may result from its heightened involvement in extracting visual object semantics.


\begin{figure}[!tp]
\vspace{7pt}
    \centering
    \begin{minipage}[c]{0.485\linewidth}
        \centering
        \includegraphics[width=\linewidth]{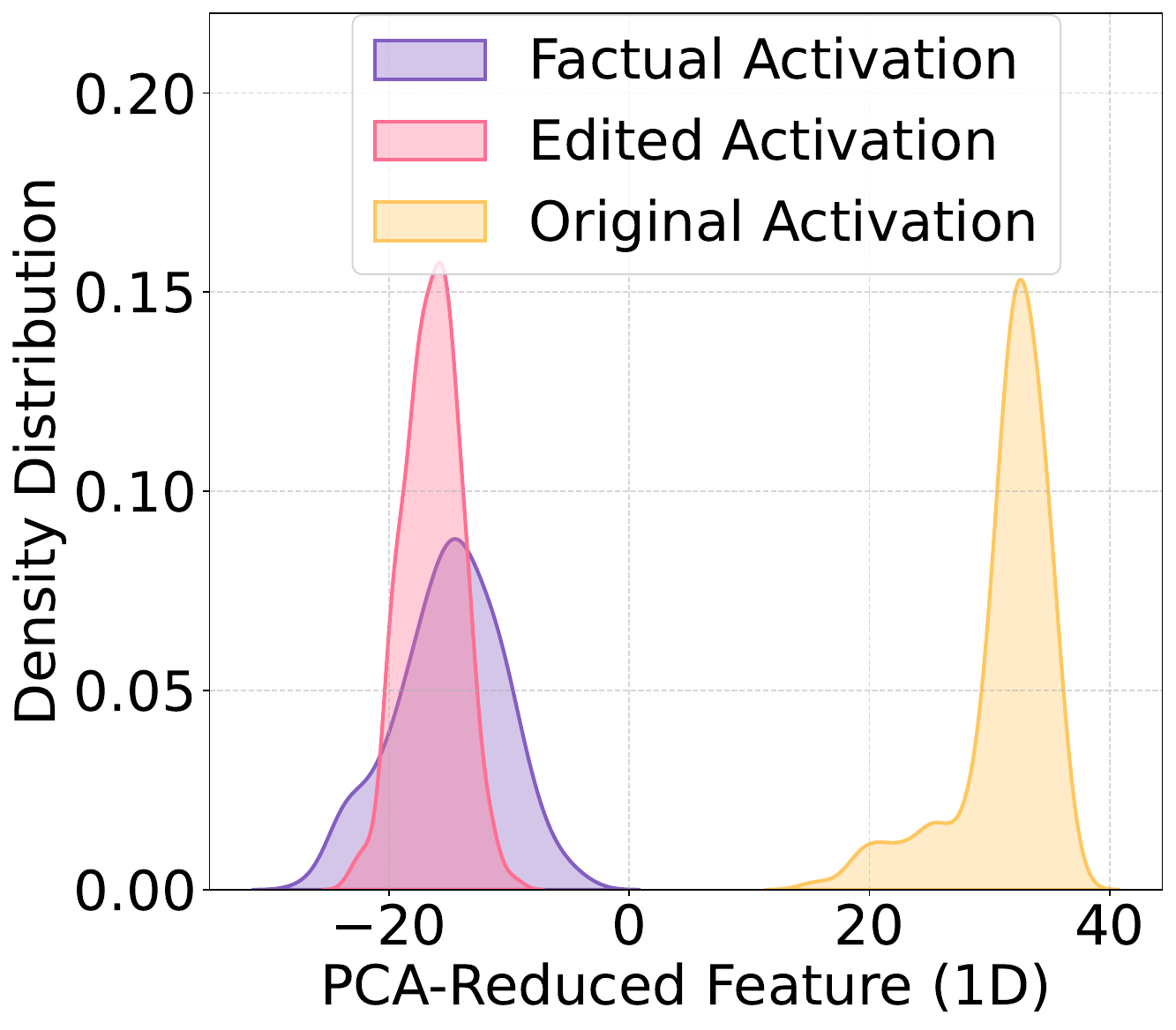}
    \end{minipage}
    \hfill
    \begin{minipage}[c]{0.48\linewidth}
        \centering
        \includegraphics[width=\linewidth]{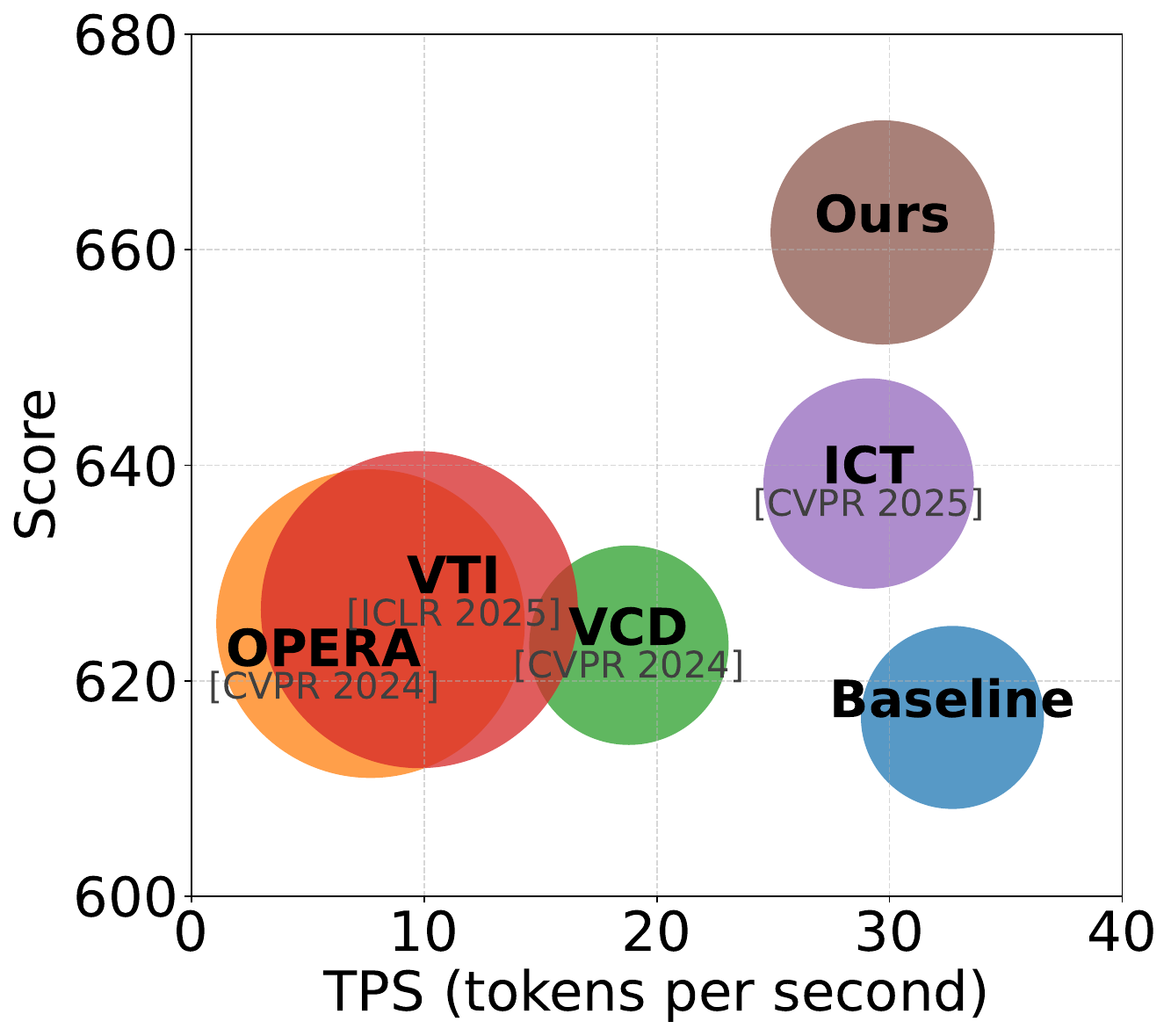}
    \end{minipage}%
    \captionof{figure}{Deep analysis on LLaVA-v1.5. \textbf{Left}: Visualization of distinct activations yielded by the last layer. \textbf{Right}: Comparison of inference speed and hallucination mitigation.}
    \label{fig:analysis_all_2}
\end{figure}

\paragraph{Visualization of Activations}\label{sec:analysis_vis}
To qualitatively investigate the mechanism of AFTER, we visualize the distributions of the last layer's factual textual activations, along with original and post-edited activations via one-dimensional PCA projections in the left of Figure \ref{fig:analysis_all_2}. It is evident that the original visual activations exhibit significant divergence from the factual textual activation distribution, highlighting the initial visual-textual disparity that leads to hallucination. After applying adaptive factual-guided visual-textual editing, the visual activations shift notably towards the textual cluster, providing evidence that AFTER indeed offers effective guidance to steering visual activations towards factual textual semantics, achieving successful mitigation of language bias.



\paragraph{Inference Computation}\label{sec:analysis_time}

We also compare inference speed and hallucination mitigation results on MME against other inference-time methods. Results in the right of Figure \ref{fig:analysis_all_2} demonstrate that our AFTER achieves the best hallucination mitigation performance while maintaining the fastest inference speed of 29.7 tokens per second. In addition, AFTER maintains moderate memory usage of 16.3 GB (expressed as the volume of spheres), facilitating practical deployment without demanding excessive resources. 



\section{Conclusion and Future Work}\label{sec:con}
In this paper, we propose AFTER, an effective activation editing approach that adaptively steers visual activation toward factual-augmented textual semantics for hallucination mitigation. Extensive experiments on typical hallucination benchmarks across three widely adopted LVLMs have confirmed that our AFTER achieves superior mitigation performance with minimal cost. It also exhibits strong generalizability and preserves general visual capabilities.  A limitation of AFTER is its dependence on the accessible activations from open-source LLMs, which restricts its applicability to closed-source LLMs. Additionally, for tasks requiring substantial domain expertise, such as medical report analysis, AFTER necessitates supplementary domain-specific data to enhance LVLM’s specialized visual perception and better mitigate language bias. In future work, we intend to extend AFTER to encompass a wider range of specialized domains.

\bibliography{aaai2026}

@inproceedings{hacl,
  title={Hallucination augmented contrastive learning for multimodal large language model},
  author={Jiang, Chaoya and Xu, Haiyang and Dong, Mengfan and Chen, Jiaxing and Ye, Wei and Yan, Ming and Ye, Qinghao and Zhang, Ji and Huang, Fei and Zhang, Shikun},
  booktitle={Proceedings of the IEEE/CVF Conference on Computer Vision and Pattern Recognition},
  pages={27036--27046},
  year={2024}
}

@inproceedings{coco,
  title={Microsoft coco: Common objects in context},
  author={Lin, Tsung-Yi and Maire, Michael and Belongie, Serge and Hays, James and Perona, Pietro and Ramanan, Deva and Doll{\'a}r, Piotr and Zitnick, C Lawrence},
  booktitle={Computer vision--ECCV 2014: 13th European conference, zurich, Switzerland, September 6-12, 2014, proceedings, part v 13},
  pages={740--755},
  year={2014},
  organization={Springer}
}

@inproceedings{pope,
  title={Evaluating Object Hallucination in Large Vision-Language Models},
  author={Li, Yifan and Du, Yifan and Zhou, Kun and Wang, Jinpeng and Zhao, Wayne Xin and Wen, Ji-Rong},
  booktitle={Proceedings of the 2023 Conference on Empirical Methods in Natural Language Processing},
  pages={292--305},
  year={2023}
}

@article{llava,
  title={Visual instruction tuning},
  author={Liu, Haotian and Li, Chunyuan and Wu, Qingyang and Lee, Yong Jae},
  journal={Advances in neural information processing systems},
  volume={36},
  pages={34892--34916},
  year={2023}
}

@inproceedings{
instructblip,
title={Instruct{BLIP}: Towards General-purpose Vision-Language Models with Instruction Tuning},
author={Wenliang Dai and Junnan Li and Dongxu Li and Anthony Tiong and Junqi Zhao and Weisheng Wang and Boyang Li and Pascale Fung and Steven Hoi},
booktitle={Thirty-seventh Conference on Neural Information Processing Systems},
year={2023},
url={https://openreview.net/forum?id=vvoWPYqZJA}
}

@article{qwen,
  title={Qwen-vl: A frontier large vision-language model with versatile abilities},
  author={Bai, Jinze and Bai, Shuai and Yang, Shusheng and Wang, Shijie and Tan, Sinan and Wang, Peng and Lin, Junyang and Zhou, Chang and Zhou, Jingren},
  journal={arXiv preprint arXiv:2308.12966},
  volume={1},
  number={2},
  pages={3},
  year={2023}
}

@article{shikra,
  title={Shikra: Unleashing multimodal llm's referential dialogue magic},
  author={Chen, Keqin and Zhang, Zhao and Zeng, Weili and Zhang, Richong and Zhu, Feng and Zhao, Rui},
  journal={arXiv preprint arXiv:2306.15195},
  year={2023}
}

@inproceedings{mplug,
  title={mplug-owl2: Revolutionizing multi-modal large language model with modality collaboration},
  author={Ye, Qinghao and Xu, Haiyang and Ye, Jiabo and Yan, Ming and Hu, Anwen and Liu, Haowei and Qian, Qi and Zhang, Ji and Huang, Fei},
  booktitle={Proceedings of the ieee/cvf conference on computer vision and pattern recognition},
  pages={13040--13051},
  year={2024}
}

@inproceedings{flickr30k,
  title={Flickr30k entities: Collecting region-to-phrase correspondences for richer image-to-sentence models},
  author={Plummer, Bryan A and Wang, Liwei and Cervantes, Chris M and Caicedo, Juan C and Hockenmaier, Julia and Lazebnik, Svetlana},
  booktitle={Proceedings of the IEEE international conference on computer vision},
  pages={2641--2649},
  year={2015}
}

@inproceedings{aokvqa,
  title={A-okvqa: A benchmark for visual question answering using world knowledge},
  author={Schwenk, Dustin and Khandelwal, Apoorv and Clark, Christopher and Marino, Kenneth and Mottaghi, Roozbeh},
  booktitle={European conference on computer vision},
  pages={146--162},
  year={2022},
  organization={Springer}
}

@article{cococaption,
  title={Microsoft coco captions: Data collection and evaluation server},
  author={Chen, Xinlei and Fang, Hao and Lin, Tsung-Yi and Vedantam, Ramakrishna and Gupta, Saurabh and Doll{\'a}r, Piotr and Zitnick, C Lawrence},
  journal={arXiv preprint arXiv:1504.00325},
  year={2015}
}

@inproceedings{gqa,
  title={Gqa: A new dataset for real-world visual reasoning and compositional question answering},
  author={Hudson, Drew A and Manning, Christopher D},
  booktitle={Proceedings of the IEEE/CVF conference on computer vision and pattern recognition},
  pages={6700--6709},
  year={2019}
}

@article{survey1,
  title={Hallucination of multimodal large language models: A survey},
  author={Bai, Zechen and Wang, Pichao and Xiao, Tianjun and He, Tong and Han, Zongbo and Zhang, Zheng and Shou, Mike Zheng},
  journal={arXiv preprint arXiv:2404.18930},
  year={2024}
}

@article{survey2,
  title={A survey on hallucination in large vision-language models},
  author={Liu, Hanchao and Xue, Wenyuan and Chen, Yifei and Chen, Dapeng and Zhao, Xiutian and Wang, Ke and Hou, Liping and Li, Rongjun and Peng, Wei},
  journal={arXiv preprint arXiv:2402.00253},
  year={2024}
}

@article{devils,
  title={Devils in middle layers of large vision-language models: Interpreting, detecting and mitigating object hallucinations via attention lens},
  author={Jiang, Zhangqi and Chen, Junkai and Zhu, Beier and Luo, Tingjin and Shen, Yankun and Yang, Xu},
  journal={arXiv preprint arXiv:2411.16724},
  year={2024}
}

@inproceedings{volcano,
  title={Volcano: Mitigating Multimodal Hallucination through Self-Feedback Guided Revision},
  author={Lee, Seongyun and Park, Sue and Jo, Yongrae and Seo, Minjoon},
  booktitle={Proceedings of the 2024 Conference of the North American Chapter of the Association for Computational Linguistics: Human Language Technologies (Volume 1: Long Papers)},
  pages={391--404},
  year={2024}
}

@inproceedings{vcd,
  title={Mitigating object hallucinations in large vision-language models through visual contrastive decoding},
  author={Leng, Sicong and Zhang, Hang and Chen, Guanzheng and Li, Xin and Lu, Shijian and Miao, Chunyan and Bing, Lidong},
  booktitle={Proceedings of the IEEE/CVF Conference on Computer Vision and Pattern Recognition},
  pages={13872--13882},
  year={2024}
}

@inproceedings{tuning,
  title={Mitigating Hallucination in Large Multi-Modal Models via Robust Instruction Tuning},
  author={Liu, Fuxiao and Lin, Kevin and Li, Linjie and Wang, Jianfeng and Yacoob, Yaser and Wang, Lijuan},
  booktitle={The Twelfth International Conference on Learning Representations},
  year={2024}
}

@inproceedings{rlhfv,
  title={Rlhf-v: Towards trustworthy mllms via behavior alignment from fine-grained correctional human feedback},
  author={Yu, Tianyu and Yao, Yuan and Zhang, Haoye and He, Taiwen and Han, Yifeng and Cui, Ganqu and Hu, Jinyi and Liu, Zhiyuan and Zheng, Hai-Tao and Sun, Maosong and others},
  booktitle={Proceedings of the IEEE/CVF Conference on Computer Vision and Pattern Recognition},
  pages={13807--13816},
  year={2024}
}

@inproceedings{dpo,
  title={Clip-dpo: Vision-language models as a source of preference for fixing hallucinations in lvlms},
  author={Ouali, Yassine and Bulat, Adrian and Martinez, Brais and Tzimiropoulos, Georgios},
  booktitle={European Conference on Computer Vision},
  pages={395--413},
  year={2024},
  organization={Springer}
}

@article{hio,
  title={Alleviating Hallucinations in Large Vision-Language Models through Hallucination-Induced Optimization},
  author={Lyu, Xinyu and Chen, Beitao and Gao, Lianli and Shen, Hengtao and Song, Jingkuan},
  journal={Advances in Neural Information Processing Systems},
  volume={37},
  pages={122811--122832},
  year={2024}
}

@inproceedings{opera,
  title={Opera: Alleviating hallucination in multi-modal large language models via over-trust penalty and retrospection-allocation},
  author={Huang, Qidong and Dong, Xiaoyi and Zhang, Pan and Wang, Bin and He, Conghui and Wang, Jiaqi and Lin, Dahua and Zhang, Weiming and Yu, Nenghai},
  booktitle={Proceedings of the IEEE/CVF Conference on Computer Vision and Pattern Recognition},
  pages={13418--13427},
  year={2024}
}

@inproceedings{halc,
  title={HALC: object hallucination reduction via adaptive focal-contrast decoding},
  author={Chen, Zhaorun and Zhao, Zhuokai and Luo, Hongyin and Yao, Huaxiu and Li, Bo and Zhou, Jiawei},
  booktitle={Proceedings of the 41st International Conference on Machine Learning},
  pages={7824--7846},
  year={2024}
}

@article{woodpecker,
  title={Woodpecker: Hallucination correction for multimodal large language models},
  author={Yin, Shukang and Fu, Chaoyou and Zhao, Sirui and Xu, Tong and Wang, Hao and Sui, Dianbo and Shen, Yunhang and Li, Ke and Sun, Xing and Chen, Enhong},
  journal={Science China Information Sciences},
  volume={67},
  number={12},
  pages={220105},
  year={2024},
  publisher={Springer}
}

@inproceedings{vti,
  title={Reducing hallucinations in large vision-language models via latent space steering},
  author={Liu, Sheng and Ye, Haotian and Zou, James},
  booktitle={The Thirteenth International Conference on Learning Representations},
  year={2024}
}

@article{ict,
  title={ICT: Image-Object Cross-Level Trusted Intervention for Mitigating Object Hallucination in Large Vision-Language Models},
  author={Chen, Junzhe and Zhang, Tianshu and Huang, Shiyu and Niu, Yuwei and Zhang, Linfeng and Wen, Lijie and Hu, Xuming},
  journal={arXiv preprint arXiv:2411.15268},
  year={2024}
}

@inproceedings{farlhf,
  title={Aligning Large Multimodal Models with Factually Augmented RLHF},
  author={Sun, Zhiqing and Shen, Sheng and Cao, Shengcao and Liu, Haotian and Li, Chunyuan and Shen, Yikang and Gan, Chuang and Gui, Liang Yan and Wang, Yu Xiong and Yang, Yiming and others},
  booktitle={Findings of the 62nd Annual Meeting of the Association for Computational Linguistics, ACL 2024},
  pages={13088--13110},
  year={2024},
  organization={Association for Computational Linguistics (ACL)}
}

@article{mme,
  title={MME: A Comprehensive Evaluation Benchmark for Multimodal Large Language Models},
  author={Fu, Chaoyou and Chen, Peixian and Shen, Yunhang and Qin, Yulei and Zhang, Mengdan and Lin, Xu and Yang, Jinrui and Zheng, Xiawu and Li, Ke and Sun, Xing and others},
  journal={arXiv preprint arXiv:2306.13394},
  year={2023}
}

@inproceedings{clip,
  title={Learning transferable visual models from natural language supervision},
  author={Radford, Alec and Kim, Jong Wook and Hallacy, Chris and Ramesh, Aditya and Goh, Gabriel and Agarwal, Sandhini and Sastry, Girish and Askell, Amanda and Mishkin, Pamela and Clark, Jack and others},
  booktitle={International conference on machine learning},
  pages={8748--8763},
  year={2021},
  organization={PmLR}
}

@article{llama,
  title={Llama: Open and efficient foundation language models},
  author={Touvron, Hugo and Lavril, Thibaut and Izacard, Gautier and Martinet, Xavier and Lachaux, Marie-Anne and Lacroix, Timoth{\'e}e and Rozi{\`e}re, Baptiste and Goyal, Naman and Hambro, Eric and Azhar, Faisal and others},
  journal={arXiv preprint arXiv:2302.13971},
  year={2023}
}

@article{vicuna,
  title={Vicuna: An open-source chatbot impressing gpt-4 with 90\%* chatgpt quality},
  author={Chiang, Wei-Lin and Li, Zhuohan and Lin, Zi and Sheng, Ying and Wu, Zhanghao and Zhang, Hao and Zheng, Lianmin and Zhuang, Siyuan and Zhuang, Yonghao and Gonzalez, Joseph E and others},
  journal={See https://vicuna. lmsys. org (accessed 14 April 2023)},
  volume={2},
  number={3},
  pages={6},
  year={2023}
}

@inproceedings{eva,
  title={Eva: Exploring the limits of masked visual representation learning at scale},
  author={Fang, Yuxin and Wang, Wen and Xie, Binhui and Sun, Quan and Wu, Ledell and Wang, Xinggang and Huang, Tiejun and Wang, Xinlong and Cao, Yue},
  booktitle={Proceedings of the IEEE/CVF conference on computer vision and pattern recognition},
  pages={19358--19369},
  year={2023}
}

@inproceedings{minigpt,
  title={MiniGPT-4: Enhancing Vision-Language Understanding with Advanced Large Language Models},
  author={Zhu, Deyao and Chen, Jun and Shen, Xiaoqian and Li, Xiang and Elhoseiny, Mohamed},
  booktitle={The Twelfth International Conference on Learning Representations},
  year={2024}
}

@article{destein,
  title={DESTEIN: Navigating Detoxification of Language Models via Universal Steering Pairs and Head-wise Activation Fusion},
  author={Li, Yu and Wei, Zhihua and Jiang, Han and Gong, Chuanyang},
  journal={arXiv preprint arXiv:2404.10464},
  year={2024}
}

@article{sea,
  title={Spectral Editing of Activations for Large Language Model Alignment},
  author={Qiu, Yifu and Zhao, Zheng and Ziser, Yftah and Korhonen, Anna and Ponti, Edoardo M and Cohen, Shay B},
  journal={Advances in Neural Information Processing Systems},
  year={2024}
}

@inproceedings{llava1.5,
  title={Improved baselines with visual instruction tuning},
  author={Liu, Haotian and Li, Chunyuan and Li, Yuheng and Lee, Yong Jae},
  booktitle={Proceedings of the IEEE/CVF Conference on Computer Vision and Pattern Recognition},
  pages={26296--26306},
  year={2024}
}

@article{vasparse,
  title={VASparse: Towards Efficient Visual Hallucination Mitigation for Large Vision-Language Model via Visual-Aware Sparsification},
  author={Zhuang, Xianwei and Zhu, Zhihong and Xie, Yuxin and Liang, Liming and Zou, Yuexian},
  journal={arXiv preprint arXiv:2501.06553},
  year={2025}
}

@inproceedings{agrawal2018don,
  title={Don't just assume; look and answer: Overcoming priors for visual question answering},
  author={Agrawal, Aishwarya and Batra, Dhruv and Parikh, Devi and Kembhavi, Aniruddha},
  booktitle={Proceedings of the IEEE conference on computer vision and pattern recognition},
  pages={4971--4980},
  year={2018}
}

@inproceedings{niu2021counterfactual,
  title={Counterfactual vqa: A cause-effect look at language bias},
  author={Niu, Yulei and Tang, Kaihua and Zhang, Hanwang and Lu, Zhiwu and Hua, Xian-Sheng and Wen, Ji-Rong},
  booktitle={Proceedings of the IEEE/CVF conference on computer vision and pattern recognition},
  pages={12700--12710},
  year={2021}
}

@inproceedings{yan2024med,
  title={Med-HVL: Automatic Medical Domain Hallucination Evaluation for Large Vision-Language Models},
  author={Yan, Qianqi and He, Xuehai and Wang, Xin Eric},
  booktitle={AAAI 2024 Spring Symposium on Clinical Foundation Models},
  year={2024}
}

@article{xie2025vlms,
  title={Are VLMs Ready for Autonomous Driving? An Empirical Study from the Reliability, Data, and Metric Perspectives},
  author={Xie, Shaoyuan and Kong, Lingdong and Dong, Yuhao and Sima, Chonghao and Zhang, Wenwei and Chen, Qi Alfred and Liu, Ziwei and Pan, Liang},
  journal={arXiv preprint arXiv:2501.04003},
  year={2025}
}

@article{jiang2024devils,
  title={Devils in middle layers of large vision-language models: Interpreting, detecting and mitigating object hallucinations via attention lens},
  author={Jiang, Zhangqi and Chen, Junkai and Zhu, Beier and Luo, Tingjin and Shen, Yankun and Yang, Xu},
  journal={arXiv preprint arXiv:2411.16724},
  year={2024}
}

@article{iti,
  title={Inference-time intervention: Eliciting truthful answers from a language model},
  author={Li, Kenneth and Patel, Oam and Vi{\'e}gas, Fernanda and Pfister, Hanspeter and Wattenberg, Martin},
  journal={Advances in Neural Information Processing Systems},
  volume={36},
  year={2024}
}

@inproceedings{trfr,
  title={Truth forest: Toward multi-scale truthfulness in large language models through intervention without tuning},
  author={Chen, Zhongzhi and Sun, Xingwu and Jiao, Xianfeng and Lian, Fengzong and Kang, Zhanhui and Wang, Di and Xu, Chengzhong},
  booktitle={Proceedings of the AAAI Conference on Artificial Intelligence},
  volume={38},
  number={19},
  pages={20967--20974},
  year={2024}
}

@inproceedings{truthx,
    title={TruthX: Alleviating Hallucinations by Editing Large Language Models in Truthful Space},
    author={Zhang, Shaolei and Yu, Tian and Feng, Yang},
    booktitle = {Proceedings of the 62nd Annual Meeting of the Association for Computational Linguistics (Volume 1: Long Papers)},
    year = {2024},
    pages = {8908--8949}
}

@inproceedings{wang2024mitigating,
  title={Mitigating fine-grained hallucination by fine-tuning large vision-language models with caption rewrites},
  author={Wang, Lei and He, Jiabang and Li, Shenshen and Liu, Ning and Lim, Ee-Peng},
  booktitle={International Conference on Multimedia Modeling},
  pages={32--45},
  year={2024},
  organization={Springer}
}

@inproceedings{kimvacode,
  title={VACoDe: Visual Augmented Contrastive Decoding},
  author={Kim, Sihyeon and Cho, Boryeong and Bae, Sangmin and Ahn, Sumyeong and Yun, Se-Young},
  booktitle={Trustworthy Multi-modal Foundation Models and AI Agents (TiFA)}
}

@article{wang2023amber,
  title={Amber: An llm-free multi-dimensional benchmark for mllms hallucination evaluation},
  author={Wang, Junyang and Wang, Yuhang and Xu, Guohai and Zhang, Jing and Gu, Yukai and Jia, Haitao and Wang, Jiaqi and Xu, Haiyang and Yan, Ming and Zhang, Ji and others},
  journal={arXiv preprint arXiv:2311.07397},
  year={2023}
}

@inproceedings{chair,
  title={Object Hallucination in Image Captioning},
  author={Rohrbach, Anna and Hendricks, Lisa Anne and Burns, Kaylee and Darrell, Trevor and Saenko, Kate},
  booktitle={Proceedings of the 2018 Conference on Empirical Methods in Natural Language Processing},
  pages={4035--4045},
  year={2018}
}
\makeatletter
\@ifundefined{isChecklistMainFile}{
  \newif\ifreproStandalone
  \reproStandalonetrue
}{
  \newif\ifreproStandalone
  \reproStandalonefalse
}
\makeatother

\ifreproStandalone
\documentclass[letterpaper]{article}
\usepackage[submission]{aaai2026}
\setlength{\pdfpagewidth}{8.5in}
\setlength{\pdfpageheight}{11in}
\usepackage{times}
\usepackage{helvet}
\usepackage{courier}
\usepackage{xcolor}
\frenchspacing

\begin{document}
\fi
\setlength{\leftmargini}{20pt}
\makeatletter\def\@listi{\leftmargin\leftmargini \topsep .5em \parsep .5em \itemsep .5em}
\def\@listii{\leftmargin\leftmarginii \labelwidth\leftmarginii \advance\labelwidth-\labelsep \topsep .4em \parsep .4em \itemsep .4em}
\def\@listiii{\leftmargin\leftmarginiii \labelwidth\leftmarginiii \advance\labelwidth-\labelsep \topsep .4em \parsep .4em \itemsep .4em}\makeatother

\setcounter{secnumdepth}{0}
\renewcommand\thesubsection{\arabic{subsection}}
\renewcommand\labelenumi{\thesubsection.\arabic{enumi}}

\newcounter{checksubsection}
\newcounter{checkitem}[checksubsection]

\newcommand{\checksubsection}[1]{%
  \refstepcounter{checksubsection}%
  \paragraph{\arabic{checksubsection}. #1}%
  \setcounter{checkitem}{0}%
}

\newcommand{\checkitem}{%
  \refstepcounter{checkitem}%
  \item[\arabic{checksubsection}.\arabic{checkitem}.]%
}
\newcommand{\question}[2]{\normalcolor\checkitem #1 #2 \color{blue}}
\newcommand{\ifyespoints}[1]{\makebox[0pt][l]{\hspace{-15pt}\normalcolor #1}}

\section*{Reproducibility Checklist}

\vspace{1em}
\hrule
\vspace{1em}

\textbf{Instructions for Authors:}

This document outlines key aspects for assessing reproducibility. Please provide your input by editing this \texttt{.tex} file directly.

For each question (that applies), replace the ``Type your response here'' text with your answer.

\vspace{1em}
\noindent
\textbf{Example:} If a question appears as
\begin{center}
\noindent
\begin{minipage}{.9\linewidth}
\ttfamily\raggedright
\string\question \{Proofs of all novel claims are included\} \{(yes/partial/no)\} \\
Type your response here
\end{minipage}
\end{center}
you would change it to:
\begin{center}
\noindent
\begin{minipage}{.9\linewidth}
\ttfamily\raggedright
\string\question \{Proofs of all novel claims are included\} \{(yes/partial/no)\} \\
yes
\end{minipage}
\end{center}
Please make sure to:
\begin{itemize}\setlength{\itemsep}{.1em}
\item Replace ONLY the ``Type your response here'' text and nothing else.
\item Use one of the options listed for that question (e.g., \textbf{yes}, \textbf{no}, \textbf{partial}, or \textbf{NA}).
\item \textbf{Not} modify any other part of the \texttt{\string\question} command or any other lines in this document.\\
\end{itemize}

You can \texttt{\string\input} this .tex file right before \texttt{\string\end\{document\}} of your main file or compile it as a stand-alone document. Check the instructions on your conference's website to see if you will be asked to provide this checklist with your paper or separately.

\vspace{1em}
\hrule
\vspace{1em}


\checksubsection{General Paper Structure}
\begin{itemize}

\question{Includes a conceptual outline and/or pseudocode description of AI methods introduced}{(yes/partial/no/NA)}
yes

\question{Clearly delineates statements that are opinions, hypothesis, and speculation from objective facts and results}{(yes/no)}
yes

\question{Provides well-marked pedagogical references for less-familiar readers to gain background necessary to replicate the paper}{(yes/no)}
yes

\end{itemize}
\checksubsection{Theoretical Contributions}
\begin{itemize}

\question{Does this paper make theoretical contributions?}{(yes/no)}
no

	\ifyespoints{\vspace{1.2em}If yes, please address the following points:}
        \begin{itemize}
	
	\question{All assumptions and restrictions are stated clearly and formally}{(yes/partial/no)}
	Type your response here

	\question{All novel claims are stated formally (e.g., in theorem statements)}{(yes/partial/no)}
	Type your response here

	\question{Proofs of all novel claims are included}{(yes/partial/no)}
	Type your response here

	\question{Proof sketches or intuitions are given for complex and/or novel results}{(yes/partial/no)}
	Type your response here

	\question{Appropriate citations to theoretical tools used are given}{(yes/partial/no)}
	Type your response here

	\question{All theoretical claims are demonstrated empirically to hold}{(yes/partial/no/NA)}
	Type your response here

	\question{All experimental code used to eliminate or disprove claims is included}{(yes/no/NA)}
	Type your response here
	
	\end{itemize}
\end{itemize}

\checksubsection{Dataset Usage}
\begin{itemize}

\question{Does this paper rely on one or more datasets?}{(yes/no)}
yes

\ifyespoints{If yes, please address the following points:}
\begin{itemize}

	\question{A motivation is given for why the experiments are conducted on the selected datasets}{(yes/partial/no/NA)}
	yes

	\question{All novel datasets introduced in this paper are included in a data appendix}{(yes/partial/no/NA)}
	yes

	\question{All novel datasets introduced in this paper will be made publicly available upon publication of the paper with a license that allows free usage for research purposes}{(yes/partial/no/NA)}
	yes

	\question{All datasets drawn from the existing literature (potentially including authors' own previously published work) are accompanied by appropriate citations}{(yes/no/NA)}
	yes

	\question{All datasets drawn from the existing literature (potentially including authors' own previously published work) are publicly available}{(yes/partial/no/NA)}
	yes

	\question{All datasets that are not publicly available are described in detail, with explanation why publicly available alternatives are not scientifically satisficing}{(yes/partial/no/NA)}
	yes

\end{itemize}
\end{itemize}

\checksubsection{Computational Experiments}
\begin{itemize}

\question{Does this paper include computational experiments?}{(yes/no)}
yes

\ifyespoints{If yes, please address the following points:}
\begin{itemize}

	\question{This paper states the number and range of values tried per (hyper-) parameter during development of the paper, along with the criterion used for selecting the final parameter setting}{(yes/partial/no/NA)}
	yes

	\question{Any code required for pre-processing data is included in the appendix}{(yes/partial/no)}
	yes

	\question{All source code required for conducting and analyzing the experiments is included in a code appendix}{(yes/partial/no)}
	yes

	\question{All source code required for conducting and analyzing the experiments will be made publicly available upon publication of the paper with a license that allows free usage for research purposes}{(yes/partial/no)}
	yes
        
	\question{All source code implementing new methods have comments detailing the implementation, with references to the paper where each step comes from}{(yes/partial/no)}
	yes

	\question{If an algorithm depends on randomness, then the method used for setting seeds is described in a way sufficient to allow replication of results}{(yes/partial/no/NA)}
	yes

	\question{This paper specifies the computing infrastructure used for running experiments (hardware and software), including GPU/CPU models; amount of memory; operating system; names and versions of relevant software libraries and frameworks}{(yes/partial/no)}
	yes

	\question{This paper formally describes evaluation metrics used and explains the motivation for choosing these metrics}{(yes/partial/no)}
	yes

	\question{This paper states the number of algorithm runs used to compute each reported result}{(yes/no)}
	yes

	\question{Analysis of experiments goes beyond single-dimensional summaries of performance (e.g., average; median) to include measures of variation, confidence, or other distributional information}{(yes/no)}
	no

	\question{The significance of any improvement or decrease in performance is judged using appropriate statistical tests (e.g., Wilcoxon signed-rank)}{(yes/partial/no)}
	no

	\question{This paper lists all final (hyper-)parameters used for each model/algorithm in the paper’s experiments}{(yes/partial/no/NA)}
	yes

\end{itemize}
\end{itemize}
\ifreproStandalone
\end{document}
\fi

\end{document}